\documentclass{article}

\usepackage{microtype}
\usepackage{graphicx}
\usepackage{subcaption}
\usepackage{booktabs}
\usepackage{multirow}

\usepackage{hyperref}

\usepackage[preprint]{icml2026}

\usepackage{amsmath}
\usepackage{amssymb}

\usepackage{newfloat}
\usepackage{listings}
\usepackage{array}
\usepackage[table]{xcolor}
\usepackage{tabularx}
\DeclareFloatingEnvironment{listing}
\floatname{listing}{Listing}
\newcolumntype{Z}{>{\centering\arraybackslash}p{1cm}}

\icmltitlerunning{CLAA: Cross-Layer Attention Aggregation for Accelerating LLM Prefill}

\begin{document}

\twocolumn[
  \icmltitle{CLAA: Cross-Layer Attention Aggregation for Accelerating LLM Prefill}

  \begin{icmlauthorlist}
    \icmlauthor{Bradley McDanel}{fm,rl}
    \icmlauthor{Steven Li}{rl}
    \icmlauthor{Harshit Khaitan}{rl}
  \end{icmlauthorlist}

  \icmlaffiliation{fm}{Franklin and Marshall College}
  \icmlaffiliation{rl}{Meta Reality Labs}
  \icmlcorrespondingauthor{Bradley McDanel}{bmcdanel@fandm.edu}
  
  \icmlkeywords{Large Language Models, Inference Acceleration, Attention Mechanisms}

  \vskip 0.3in
]

\printAffiliationsAndNotice{}

\begin{abstract}
The prefill stage in long-context LLM inference remains a computational bottleneck. Recent token-ranking heuristics accelerate inference by selectively processing a subset of semantically relevant tokens. However, existing methods suffer from unstable token importance estimation, often varying between layers. Evaluating token-ranking quality independently from heuristic-specific architectures is challenging. To address this, we introduce an Answer-Informed Oracle, which defines ground-truth token importance by measuring attention from generated answers back to the prompt. This oracle reveals that existing heuristics exhibit high variance across layers: rankings can degrade sharply at specific layers, a failure mode invisible to end-to-end benchmarks. The diagnosis suggests a simple fix: aggregate scores across layers rather than relying on any single one. We implement this as Cross-Layer Attention Aggregation (CLAA), which closes the gap to the oracle upper bound and reduces Time-to-First-Token (TTFT) by up to 39\% compared to the Full KV Cache baseline.
\end{abstract}

\section{Introduction}
\label{sec:intro}

\begin{figure}[t]
    \centering
    \includegraphics[width=\linewidth]{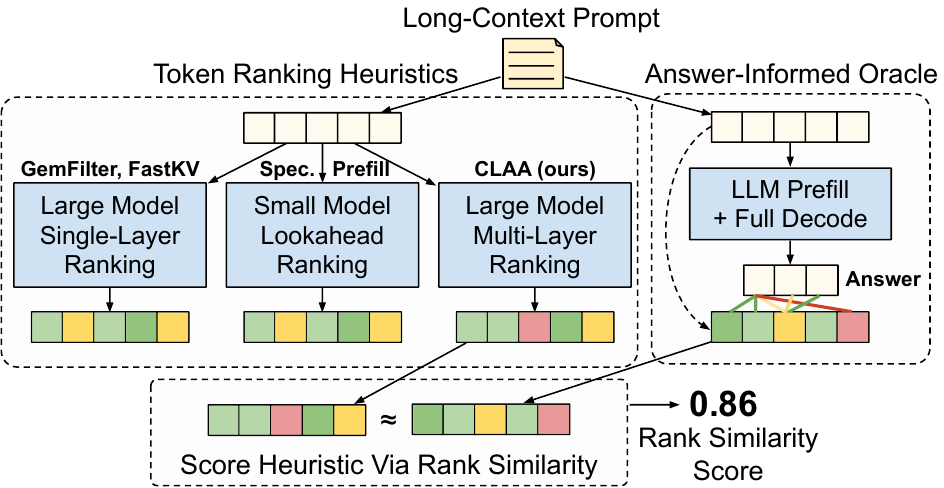}
    \caption{Illustration of our framework for evaluating token ranking heuristics for LLM Prefill acceleration. An Answer-Informed Oracle establishes a ground-truth token ranking by aggregating attention from the generated answer back to the prompt. This approach, which measures rank similarity between heuristic outputs and the oracle, motivates our Cross-Layer Attention Aggregation (CLAA) method that achieves higher alignment with the oracle.}
    \label{fig:overview}
\end{figure}

One emerging framework to reduce this high computational cost is to operate on a smaller subset of only the most important prompt tokens. For instance, when processing a long document, many of the tokens can be irrelevant to the posed query.

The specific heuristics employed within this framework vary in both their ranking signals and their architectural designs. For ranking, importance scores are derived from different sources. Methods like GemFilter~\cite{shi2024discovering} and FastKV~\cite{jo2025fastkv} derive their ranking signal from the final tokens of the input prompt. In contrast, Speculative Prefill~\cite{liu2025speculative} uses a separate smaller model to draft a plausible future continuation, which then serves as the basis for scoring the prompt. Architecturally, these approaches can differ, with some restarting computation entirely, while others dynamically prune tokens during the forward pass. The variety of approaches makes principled comparison difficult. Current benchmarks rely on a single end-to-end performance metric, which conflates ranking quality with architectural efficiency and obscures how token importance evolves across different layers and attention heads.

To address these challenges, we propose an Answer-Informed Oracle (Figure~\ref{fig:overview}) that establishes a ground-truth for evaluating token ranking heuristics. Our approach is founded on the observation that the true importance of a prompt token is determined by the attention it receives from the generated answer.

Using this framework, we find that aggregating attention scores across multiple model layers produces a more robust signal of token importance. This analysis suggests a straightforward fix: aggregate importance scores across layers rather than trusting any single one. We implement this as Cross-Layer Attention Aggregation (CLAA), which produces rankings with higher similarity to the oracle and validates that the identified instability was indeed the bottleneck. In summary, our main contributions are:
\begin{itemize}
    \item We introduce an \textbf{Answer-Informed Oracle} that establishes ground-truth token importance by measuring attention from the generated answer back to the prompt.

    \item Using this oracle, we develop a \textbf{quantitative evaluation framework} that reveals layer-wise ranking instability in existing heuristics, a failure mode invisible to end-to-end benchmarks.

    \item Guided by this diagnosis, we implement \textbf{Cross-Layer Attention Aggregation} (CLAA), a simple aggregation strategy that validates our analysis by closing the gap to the oracle.
\end{itemize}

\section{Related Work}
\label{sec:related}

\subsection{Approaches to Prefill Acceleration}
\label{sec:related:prefill}

Hardware-aware optimizations such as FlashAttention~\cite{dao2022flashattention,dao2023flashattention} achieve substantial speedups through improved memory access patterns, though they still maintain quadratic complexity with respect to sequence length. Researchers have developed dynamic sparse attention methods to address this fundamental scaling limitation. Methods like SpAtten~\cite{wang2021spatten}, H2O~\cite{zhang2023h2o}, and MInference~\cite{jiang2024minference} use the observation that attention patterns vary dynamically with input content, adaptively identifying important tokens at runtime rather than relying on static sparsity patterns.

Alternative architectural solutions offer different trade-offs. State space models like Mamba~\cite{gu2023mamba} achieve linear complexity but face challenges with precise token recall, which has led to the development of hybrid architectures such as Jamba~\cite{lieber2024jamba}. Semantic compression techniques including LLMLingua~\cite{jiang2023llmlingua,pan2024llmlingua} employ smaller auxiliary models to rewrite lengthy prompts into more concise representations, achieving substantial compression ratios but introducing computational overhead from the compression process itself. Architectures like YOCO~\cite{sun2024you} require additional training or fine-tuning phases.

Token-ranking heuristics take a different approach. These methods are training-free and reduce computational costs by identifying and processing only the most relevant token subsets. They can be applied directly to existing models without modification. This paper presents an evaluation framework and improved heuristic for such approaches.

\subsection{Token Ranking Strategies in Prefill Acceleration}
\label{sec:related:ranking}

Several recent methods accelerate the prefill stage by avoiding a full-sequence forward pass through all layers. These approaches share a common strategy of first ranking the importance of prompt tokens using an initial, less computationally expensive ranking heuristic. This ranking allows the model to perform its main, compute-intensive forward pass on only a top-ranked subset of tokens, thereby lowering the total prefill cost. While these heuristics all rely on attention scores to perform this ranking, their primary distinction lies in the choice of query vectors (Q) used to probe the key vectors (K) of the prompt. To clarify these differences, we formalize each approach below using a consistent notation.

Let $T_{\text{prompt}}$ be an input prompt of length $L$. For a given model $M$ and layer $l$, let $K_{\text{prompt}}^{(l,h)} \in \mathbb{R}^{L \times d_k}$ be the matrix of key vectors for all prompt tokens at head $h$, where $d_k$ is the head dimensionality. The goal of each strategy is to compute an importance score $S_i$ for each prompt token $i \in \{1, \dots, L\}$.

\paragraph{GemFilter.} GemFilter \cite{shi2024discovering} hypothesizes that the query from the \textit{very last token} of the prompt, after being processed by some initial layers, is sufficient to identify relevant context. It runs the model $M$ for $r$ layers to produce the query vector for the last token, $\mathbf{q}_{\text{last}}^{(r)}$. The importance score for the $i$-th prompt token is then computed by summing the raw attention scores (pre-softmax) from this single query across all attention heads $h$:
\begin{equation}
    S_i^{\text{GF}} = \sum_{h} \left[ \frac{\mathbf{q}_{\text{last}}^{(r, h)} (K_{\text{prompt}}^{(r, h)})^\top}{\sqrt{d_k}} \right]_i
\end{equation}

\paragraph{FastKV.} FastKV \cite{jo2025fastkv} uses a small observation window, $\mathcal{W}$, consisting of the $W$ most recent prompt tokens as queries. This allows a collective assessment from multiple positions at the end of the context. These queries are selected from a specific Token-Selective Propagation (TSP) layer, denoted $l_{\text{TSP}}$. The importance score is derived by summing the post-softmax attention probabilities from each query in the window across all attention heads $h$:
\begin{equation}
    S_i^{\text{FKV}} = \sum_{j \in \mathcal{W}} \sum_{h} \left[ \text{Softmax}\left(\frac{\mathbf{q}_{j}^{(l_{\text{TSP}}, h)} (K_{\text{prompt}}^{(l_{\text{TSP}}, h)})^\top}{\sqrt{d_k}}\right) \right]_i
\end{equation}

\paragraph{Speculative Prefill.}
In contrast, Speculative Prefill \cite{liu2025speculative} uses a separate, smaller \textit{speculator model}, $M_{\text{spec}}$, to look into the ``future.'' It generates $k$ lookahead tokens and uses their corresponding query vectors, $\left\{ \mathbf{q}_{\text{gen}, j} \right\}_{j=1}^{k}$, to score the prompt. This assesses token importance based on what a model \textit{would} look for while generating a plausible continuation. The final score is the mean of the maximum raw attention scores (pre-softmax) from each lookahead query, taken across all layers and heads:
\begin{equation}
    S_i^{\text{SP}} = \frac{1}{k} \sum_{j=1}^{k} \left( \operatorname*{max}_{l, h} \left[ \frac{\mathbf{q}_{\text{gen}, j}^{(l, h)} (K_{\text{prompt, spec}}^{(l, h)})^\top}{\sqrt{d_k}} \right]_i \right)
\end{equation}
where $K_{\text{prompt, spec}}^{(l, h)}$ are the prompt key vectors as computed by the speculator model $M_{\text{spec}}$.

\subsection{KV Cache Management in Prefill Acceleration}
\label{sec:related:kv_management}

In addition to pruning tokens during prefill, ranking heuristics are also used to compress the KV cache for decode.

\paragraph{Compression via Sequence Pruning.}
For methods like GemFilter and Speculative Prefill, KV cache compression is a direct consequence of their architectural design. These heuristics first identify a single, shared subset of $L_{\text{pruned}}$ important tokens. They then execute a second, main forward pass using only this pruned sequence. The compression follows naturally: the KV cache is built exclusively for these $L_{\text{pruned}}$ tokens. For any given layer $l$, the resulting key cache matrix, $K^{(l)} \in \mathbb{R}^{L_{\text{pruned}} \times d_k}$, is therefore uniform across all attention heads.

\paragraph{Layer-wise Cache Compression.}
In contrast, FastKV employs a dual-strategy architecture that performs explicit compression of the KV cache at each layer. At each layer $l$ before the final pruning step, it computes an importance score, $\mathcal{I}$, for each key-value head group $g$:
\begin{equation}
    \mathcal{I}_{i,g}^{\text{KV-FKV}} = \frac{1}{|\mathcal{H}_g|} \sum_{h \in \mathcal{H}_g} \sum_{j \in \mathcal{W}} \left[ \text{Softmax}\left(\frac{\mathbf{q}_{j}^{(l, h)} (K_{\text{prompt}}^{(l, g)})^\top}{\sqrt{d_k}}\right) \right]_i
\end{equation}
Based on these scores, a compressed KV cache is stored independently for each head group. Critically, while the stored cache is compressed, the full, unpruned hidden states are propagated to the subsequent layer for computation. This separation allows full-context processing with a memory-efficient cache.

\section{Answer-Informed Oracle Framework}
\label{sec:oracle}

\subsection{Oracle Construction}
\label{sec:oracle:construction}

Our oracle is based on the idea that prompt token importance is definitively measured by attention received from the generated answer. Unlike existing heuristics predicting future importance from partial information, our oracle uses full knowledge of the generation process to determine precisely which prompt tokens influenced the response.

Algorithm~\ref{alg:oracle-ranking} details the three-stage process for constructing oracle rankings. First, we process the full prompt through the model to extract key vectors for all prompt tokens. Second, we generate a complete answer while capturing the query vectors from each generated token. This separation into stages allows us to compute the exact attention scores we need without incurring the prohibitive memory cost of storing the full attention matrices from every generation step. Finally, we compute attention scores between all answer queries and prompt keys, aggregating these scores to produce a definitive importance ranking for each prompt token.

The oracle design incorporates several architectural choices. We adopt a maximum aggregation strategy across layers and attention heads, similar to that used in Speculative Prefill \cite{liu2025speculative}, to capture the peak importance of each token. This approach recognizes that even a single strong attention connection can indicate critical information. However, whereas Speculative Prefill aggregates attention from a handful of speculated future tokens, our oracle averages these peak scores across the entire ground-truth generated answer, providing a measure of each token’s true importance. Finally, we apply 1D average pooling with a small kernel to reduce local variations and stabilize rankings.

\begin{algorithm}[t]
\small
\caption{Answer-Informed Oracle Token Ranking}
\label{alg:oracle-ranking}
\begin{algorithmic}[1]
\REQUIRE Oracle Model $M$, prompt tokens $T_{\text{prompt}}$, max output $N_{\text{gen}}$
\STATE $S_{\text{prompt}} \gets \textsc{ModelForward}(M, T_{\text{prompt}})$
\STATE $K_{\text{prompt}} \gets \textsc{GetKeys}(S_{\text{prompt}})$ \hfill $\triangleright$ Extract keys from prompt
\STATE Initialize empty lists $Q_{\text{gen}}, T_{\text{gen}}$
\FOR{$i = 1$ to $N_{\text{gen}}$}
    \STATE $S \gets \textsc{ModelForward}(M, T_{\text{prompt}} \oplus T_{\text{gen}})$ \hfill $\triangleright$ Generate oracle answer
    \STATE $Q_{\text{gen}}[i] \gets \textsc{GetLastQuery}(S)$ \hfill $\triangleright$ Store query
    \STATE $t \gets \text{argmax}(\textsc{GetLastLogit}(S))$
    \IF{$t$ is EOS}
        \STATE \textbf{break}
    \ENDIF
    \STATE $T_{\text{gen}}[i] \gets t$ \hfill $\triangleright$ Store token
\ENDFOR
\STATE $A \gets \textsc{ComputeAttn}(Q_{\text{gen}}, K_{\text{prompt}})$ \hfill $\triangleright$ All attention scores
\STATE $S_{\text{oracle}} \gets \textsc{Pool1D}(\textsc{MeanAgg}(\textsc{MaxAgg}(A)))$ \hfill $\triangleright$ Aggregate \& denoise
\STATE \textbf{return} $S_{\text{oracle}}$
\end{algorithmic}
\end{algorithm}

To quantitatively assess how well a ranking generated by a heuristic aligns with the ground truth established by the oracle, we use the Spearman Rank Correlation ($\rho$). This metric is a global measure of ranking quality by comparing the ranks assigned to each token by the oracle and heuristic methods. Values range from -1 (perfect disagreement) to 1 (perfect agreement), offering a direct measure of the ordinal similarity between the heuristic and oracle rankings.

\subsection{Oracle as an Upper-Bound Benchmark}
\label{sec:oracle:upperbound}
Beyond comparing rankings via similarity metrics, the Answer-Informed Oracle is an empirical upper bound for any given token keep rate. This involves first computing the oracle ranking for a prompt using Algorithm~\ref{alg:oracle-ranking}, then executing a separate forward pass using only the top-k\% tokens identified by this ranking.

To ensure a fair upper bound, the oracle-guided forward pass must precisely mirror the architectural strategy of the evaluated heuristic. For methods like GemFilter and FastKV that prune at an intermediate layer $l_p$, the oracle emulation follows a specific protocol: in layers before $l_p$, pre-computed oracle rankings compress the KV cache by selecting top-ranked token pairs, while full hidden states propagate forward to preserve the computational path. At the pruning layer $l_p$, these same rankings perform sequence pruning, filtering the hidden states themselves to retain only top-ranked tokens. This reduced sequence then flows through all subsequent layers, with the KV cache at $l_p$ compressed using the same indices.

\section{Cross-Layer Attention Aggregation (CLAA)}
\label{sec:claa}

\subsection{The Oracle Reveals Layer-wise Instability}
\label{sec:claa:limitations}

\begin{figure}[t]
    \centering
    \includegraphics[width=\linewidth]{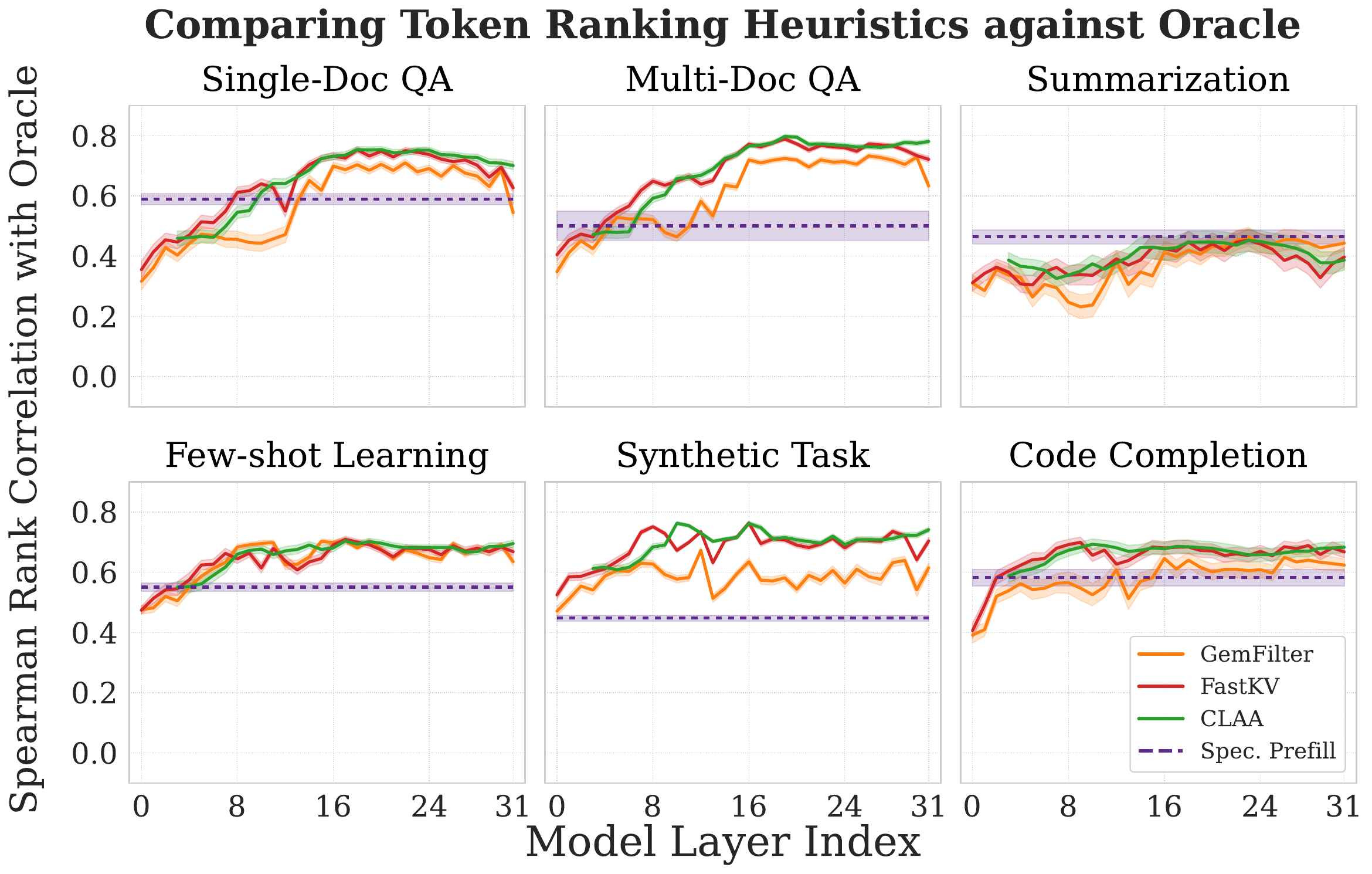}
    \caption{Layer-wise token ranking performance on Llama-3.1-8B-Instruct. Spearman correlation with answer-informed oracle across LongBench tasks, comparing existing heuristics to our proposed CLAA method.}
    \label{fig:ranking-comparison}
\end{figure}

Figure~\ref{fig:ranking-comparison} reveals two limitations of existing token-ranking methods when examined layer by layer against our oracle baseline. First, heuristics such as FastKV and GemFilter exhibit high variance, with rank correlations fluctuating dramatically and experiencing sharp drops at specific layers. This volatility is risky for methods that depend on a single layer to determine sequence pruning decisions. When the selected layer (e.g., $l_{\text{TSP}}$) coincides with one of these performance troughs, the resulting token rankings become unreliable, potentially eliminating important tokens and compromising downstream task performance.

Second, our analysis reveals that early layers consistently show unreliable token rankings. The rank correlations in initial layers (particularly layers 0-4) are lower than those in deeper layers. Pruning based on early-layer rankings uses low-fidelity signals. Consequently, essential key-value pairs may be prematurely discarded, undermining the model's ability to construct accurate semantic representations in subsequent layers and degrading the contextual information available for generation.

\subsection{Multi-Layer Aggregation for Robust Ranking}
\label{sec:claa:method}

The Oracle analysis reveals that single-layer heuristics are unstable, with performance varying sharply across layers. The implication is direct: do not trust any single layer. We implement this as Cross-Layer Attention Aggregation (CLAA), which aggregates scores across consecutive layers. First, given the unreliability of initial layers for ranking decisions, CLAA defers KV cache compression during the first $m=4$ layers of the model. By maintaining the complete context in these early stages, we ensure that the model can construct high-fidelity semantic representations before any KV compression occurs. Second, rather than relying on potentially unstable single-layer rankings, CLAA aggregates token importance signals across multiple consecutive layers. By taking the maximum importance score across these layers, tokens critical to any layer are preserved, making the method robust against layer-specific failures. At a designated pruning layer $l_p$, CLAA synthesizes information from a window of $n$ preceding layers, $\mathcal{L} = \{l_p - n + 1, \dots, l_p\}$. For each prompt token $i$ and layer $l' \in \mathcal{L}$, we compute a layer-specific importance score $S_i^{(l')}$ based on attention from a small observation window $\mathcal{W}$ containing the $W$ most recent prompt tokens:

\begin{equation}
S_i^{(l')} = \sum_{j \in \mathcal{W}} \sum_{h} \left[ \text{Softmax}\left( \frac{\mathbf{q}_j^{(l',h)} (K_{\text{prompt}}^{(l',h)})^\top}{\sqrt{d_k}} \right) \right]_i
\label{eq:claa_layer_score}
\end{equation}

Here, $\mathbf{q}_j^{(l',h)} \in \mathbb{R}^{d_k}$ represents the query vector from the $j$-th token in the observation window at layer $l'$ and head $h$, while $K_{\text{prompt}}^{(l',h)} \in \mathbb{R}^{L \times d_k}$ contains key vectors for all $L$ prompt tokens.

When the model reaches the pruning layer $l_p$, CLAA computes the final importance score for each token by taking the maximum across all collected layer scores: $S_i^{\text{CLAA}} = \max_{l' \in \mathcal{L}} S_i^{(l')}$. This maximum aggregation strategy preserves tokens deemed important by any recent layer while filtering layer-specific noise. Unlike single-layer methods, CLAA is robust against the layer-specific variations shown in Figure~\ref{fig:ranking-comparison}. By aggregating across layers, CLAA produces more stable rankings. The improvement over single-layer methods validates that instability was indeed the bottleneck diagnosed by the Oracle.

\section{Experiments}
\label{sec:experiments}

\begin{table*}[!ht]
  \centering
  \setlength{\tabcolsep}{1pt}
  \caption{LLaMA-3.1-8B LongBench results across varying token keep rate (\%).}
  \label{tab:longbench_results_generated}
  \renewcommand{\arraystretch}{1.1}
  \scalebox{0.75}{
    \begin{tabular}{l|*{3}{Z}|*{3}{Z}|*{3}{Z}|*{3}{Z}|*{2}{Z}|*{2}{Z}|Z}
      \toprule
        & \multicolumn{3}{c|}{Single-Document QA} & \multicolumn{3}{c|}{Multi-Document QA} & \multicolumn{3}{c|}{Summarization} & \multicolumn{3}{c|}{Few-shot Learning} & \multicolumn{2}{c|}{Synthetic} & \multicolumn{2}{c|}{Code} &     \\
      \cmidrule(lr){2-4} \cmidrule(lr){5-7} \cmidrule(lr){8-10} \cmidrule(lr){11-13} \cmidrule(lr){14-15} \cmidrule(lr){16-17}
      Method & \rotatebox{60}{NrtvQA} & \rotatebox{60}{Qasper} & \rotatebox{60}{MF-en} & \rotatebox{60}{HotpotQA} & \rotatebox{60}{2WikiMQA} & \rotatebox{60}{MuSiQue} & \rotatebox{60}{GovReport} & \rotatebox{60}{QMSum} & \rotatebox{60}{MultiNews} & \rotatebox{60}{TREC} & \rotatebox{60}{TriviaQA} & \rotatebox{60}{SAMSum} & \rotatebox{60}{PCount} & \rotatebox{60}{PRe} & \rotatebox{60}{LCC} & \rotatebox{60}{RB-P} & \rotatebox{60}{Avg.} \\
      \midrule
      \midrule
\multicolumn{18}{c}{\textbf{Keep Token Rate = 100\%}} \\
\midrule
FullKV    &  30.16 &  45.53 &  54.94 &  56.02 &  46.66 &  31.28
          &  35.12 &  25.28 &  27.25 &  73.00 &  91.65 &  43.80
          &   8.88 &  99.50 &  63.38 &  56.73 & \textbf{49.32} \\
\midrule
\multicolumn{18}{c}{\textbf{Keep Token Rate = 10\%}} \\
\midrule
Oracle    &  29.85 &  43.94 &  55.85 &  54.99 &  47.11 &  28.92
          &  32.26 &  25.29 &  21.95 &  68.50 &  91.43 &  41.80
          &   7.47 &  99.50 &  59.63 &  56.85 & 47.83 \\ \hline
GemFilter &  24.36 &  21.07 &  39.73 &  51.29 &  33.92 &  25.78
          &  28.94 &  18.98 &  17.42 &  60.50 &  91.53 &  40.39
          &   4.76 &  87.50 &  22.73 &  32.47 & 37.59 \\
FastKV    &  30.60 &  38.96 &  53.61 &  54.87 &  44.73 &  30.09
          &  28.08 &  24.57 &  20.93 &  70.00 &  92.38 &  42.69
          &   6.56 &  99.00 &  58.43 &  53.49 & 46.81 \\
SpecPrefill &  28.53 &  32.86 &  51.94 &  54.33 &  40.80 &  29.66
          &  27.47 &  22.43 &  19.76 &  62.50 &  89.31 &  40.14
          &   4.40 &  66.08 &  50.49 &  51.09 & 41.99 \\
\rowcolor[HTML]{E8F8E8} CLAA      &  31.09 &  42.36 &  53.68 &  53.83 &  44.73 &  31.53
          &  28.15 &  24.76 &  20.42 &  70.00 &  92.37 &  42.93
          &   6.51 &  99.50 &  58.31 &  53.86 & \textbf{47.13} \\
\midrule
\multicolumn{18}{c}{\textbf{Keep Token Rate = 20\%}} \\
\midrule
Oracle    &  29.86 &  44.37 &  55.33 &  54.57 &  47.17 &  29.83
          &  33.96 &  24.92 &  24.39 &  69.50 &  91.63 &  42.93
          &   5.41 &  99.50 &  63.08 &  57.74 & 48.39 \\ \hline
GemFilter &  25.86 &  30.74 &  47.30 &  57.95 &  43.72 &  29.64
          &  31.19 &  20.92 &  21.06 &  64.00 &  92.75 &  40.92
          &   6.18 &  97.00 &  29.17 &  38.22 & 42.29 \\
FastKV    &  30.54 &  40.45 &  54.42 &  53.67 &  47.25 &  30.12
          &  28.49 &  24.33 &  21.46 &  70.50 &  92.08 &  42.85
          &   7.09 &  99.50 &  59.68 &  54.93 & 47.33 \\
SpecPrefill &  27.41 &  40.09 &  56.67 &  56.39 &  40.05 &  27.55
          &  30.28 &  23.61 &  22.52 &  62.50 &  91.04 &  41.19
          &   3.00 &  87.50 &  52.90 &  50.49 & 44.57 \\
\rowcolor[HTML]{E8F8E8} CLAA      &  30.63 &  43.95 &  53.85 &  54.61 &  46.09 &  31.01
          &  30.72 &  25.33 &  23.14 &  71.00 &  92.08 &  43.14
          &   7.61 &  99.50 &  61.21 &  56.07 & \textbf{48.12} \\
\midrule
\multicolumn{18}{c}{\textbf{Keep Token Rate = 40\%}} \\
\midrule
Oracle    &  30.14 &  45.74 &  55.36 &  54.27 &  47.42 &  30.95
          &  35.11 &  25.17 &  25.97 &  70.00 &  91.79 &  42.56
          &   6.84 &  99.50 &  63.99 &  57.33 & 48.88 \\ \hline
GemFilter &  26.90 &  39.14 &  52.33 &  56.02 &  46.85 &  32.22
          &  33.56 &  22.80 &  24.36 &  68.50 &  91.67 &  43.84
          &   4.65 &  99.50 &  37.34 &  42.04 & 45.11 \\
FastKV    &  30.39 &  40.92 &  55.09 &  55.40 &  47.36 &  31.02
          &  28.21 &  24.39 &  22.02 &  71.00 &  91.87 &  42.89
          &   6.09 &  99.50 &  61.32 &  55.47 & 47.68 \\
SpecPrefill &  27.05 &  44.00 &  54.74 &  56.27 &  44.05 &  30.65
          &  32.67 &  24.93 &  24.88 &  67.50 &  91.30 &  42.98
          &   1.89 &  97.00 &  55.90 &  49.00 & 46.55 \\
\rowcolor[HTML]{E8F8E8} CLAA      &  30.37 &  45.84 &  54.32 &  55.02 &  47.38 &  31.72
          &  33.10 &  24.87 &  25.21 &  71.00 &  92.04 &  43.31
          &   7.18 &  99.50 &  62.40 &  56.23 & \textbf{48.72} \\
      \bottomrule
    \end{tabular}
  }
\end{table*}

\subsection{Experimental Setup}
\label{sec:experiments:setup}

\paragraph{Models and Datasets.} We evaluate our approach on Llama-3.2-3B-Instruct (28 layers)~\cite{grattafiori2024llama}, Llama-3.1-8B-Instruct (32 layers), and Mistral-Nemo-12B-Instruct (40 layers)~\cite{mistralai2024nemo}. We evaluate token ranking quality using three benchmarks: (1) LongBench~\cite{bai2024longbench} covers 16 English tasks across single/multi-document QA, summarization, few-shot learning, code completion, and synthetic tasks; (2) Needle-in-a-Haystack~\cite{kamradt2023needle} tests information retrieval by embedding facts at various depths in 16K-64K token contexts; (3) RULER~\cite{hsieh2024ruler} evaluates retrieval, multi-hop tracing, information aggregation, and QA across 12 subtasks using 64K contexts with 500 samples per subtask.

\paragraph{Baselines.} We compare CLAA against three recent token ranking heuristics. GemFilter~\cite{shi2024discovering} uses attention from the last prompt token after processing through early layers. FastKV~\cite{jo2025fastkv} employs an observation window of recent tokens to score importance. Speculative Prefill~\cite{liu2025speculative} uses Llama-3.2-1B-Instruct as a draft model to generate 8 lookahead tokens for scoring; we use a 1B draft to represent resource-constrained deployment scenarios where larger drafters are not feasible. For fair comparison, all methods run on the same HuggingFace stack with FlashAttention-2, and Speculative Prefill token rankings were verified to match the original authors' implementation. Absolute TTFT numbers differ from the original paper due to our use of HF Transformers rather than vLLM with PagedAttention; results are intended for relative comparison across methods.

To ensure an equitable comparison, a given token keep rate is applied consistently to both sequence pruning during prefill and KV cache compression for decoding across all methods. For both FastKV and CLAA, the observation window size ($W$) is set to 8 tokens. To stabilize token rankings, we apply 1D average pooling with a kernel size of 7 to the importance scores for all applicable methods. Unless otherwise specified, all layer-based ranking methods (GemFilter, FastKV, and CLAA) use layer 15 of the respective model as the pruning layer. For CLAA, we set the cross-layer aggregation window size ($n$) to 4 and keep the KV cache for the first $m=4$ layers uncompressed. These choices are justified by ablation studies: pruning layer sensitivity is analyzed in Figure~\ref{fig:ablations} (left), cross-layer window size in Figure~\ref{fig:ablations} (right), and first uncompressed layer in Appendix~\ref{app:ablation_m}. All hyperparameter configurations are summarized in Table~\ref{tab:hyperparameters}.

\paragraph{Evaluation Details.} For task evaluation, we report task-specific metrics from LongBench (F1 for QA tasks, Rouge-L for summarization, accuracy for classification) and binary retrieval accuracy for Needle-in-a-Haystack. For the RULER benchmark, we report the official recall-based accuracy averaged across its tasks. Experiments used a single A100 GPU (80GB). For the oracle, we use the same main model to generate complete answers. All settings use greedy decoding (temperature 0).

\subsection{Main Results}
\label{sec:experiments:results}

\begin{figure}[t]
\centering
\includegraphics[width=\columnwidth]{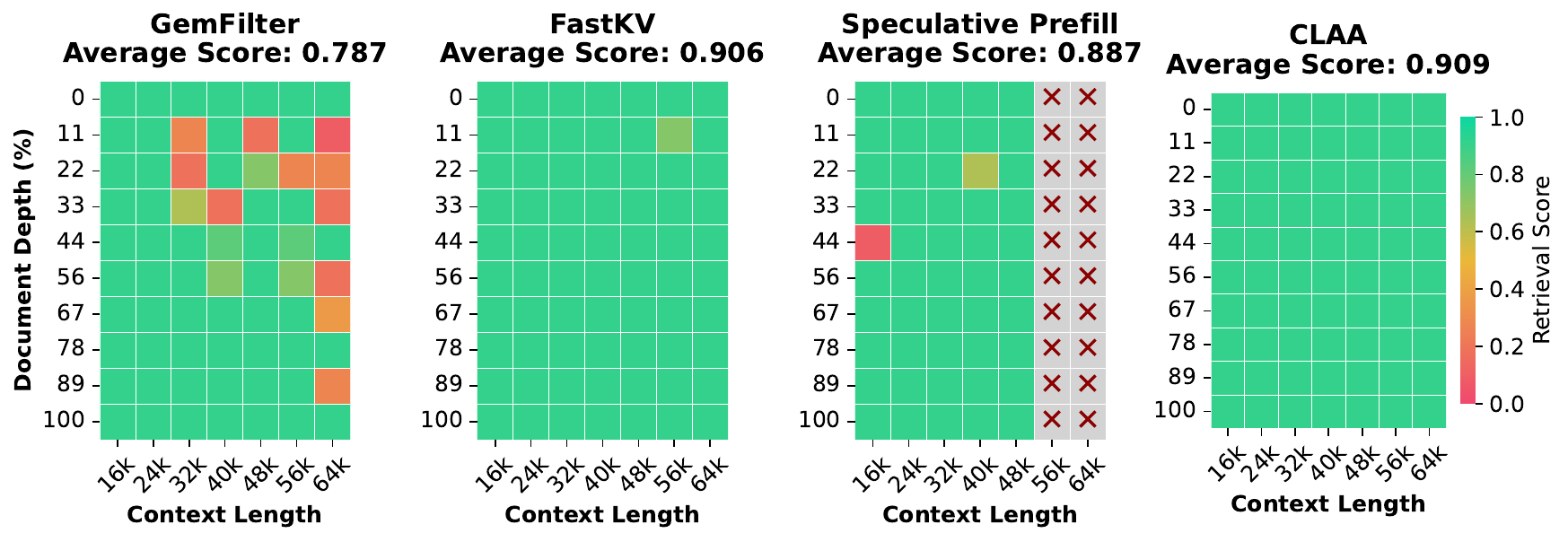}
\caption{Needle-in-a-Haystack result of LLaMA-3.1-8B-Instruct with 40\% token keep rate. X denotes out of memory on 80GB A100.}
\label{fig:needle}
\end{figure}

Our experiments validate the Oracle diagnosis. Aggregating across layers (CLAA) consistently closes the gap to the oracle ceiling, confirming that layer instability was the bottleneck. On LongBench with Llama-3.1-8B (Table~\ref{tab:longbench_results_generated}), CLAA achieves 47.13\% at 10\% token retention, within 0.7\% of the oracle ceiling (47.83\%). The gap to FastKV (46.81\%) is modest, but CLAA consistently closes more of the distance to the oracle across all keep rates. At 20\% keep rate, CLAA achieves higher accuracy than FastKV at 40\% (48.12\% vs.\ 47.68\%). The trend holds across model scales (Table~\ref{tab:longbench_avg_comparison}), confirming that multi-layer aggregation provides more reliable token selection than single-layer approaches.

CLAA also demonstrates greater robustness in challenging retrieval tasks. The Needle-in-a-Haystack results (Figure~\ref{fig:needle}) reveal critical weaknesses in baseline methods. GemFilter entirely fails to retrieve needles at intermediate document positions (22-44\% depth), whereas SpecPrefill consistently misses needles located in the latter half of documents. In contrast, CLAA maintains consistent retrieval accuracy across all needle positions, achieving the highest average score (0.909). CLAA also excels in RULER (Table~\ref{tab:ruler_results_categorized}), particularly at retrieval (89.85\%) and multi-hop reasoning (87.72\%), both tasks requiring the identification and integration of distributed information.

These retrieval results align with the Oracle analysis. Tasks that require locating specific facts distributed across context benefit from multi-layer aggregation: a token ignored at one layer may be deemed important at another. GemFilter's failure at intermediate document positions (Figure~\ref{fig:needle}) suggests its routing layer attends primarily to document boundaries, missing mid-document content. By aggregating across layers, CLAA captures tokens that any layer considers relevant, avoiding blind spots inherent to single-layer decisions.

\begin{table}[t]
  \centering
  \caption{Average LongBench results across different models and token keep rates.}
  \label{tab:longbench_avg_comparison}
  \renewcommand{\arraystretch}{1.2}
  \newcolumntype{C}{>{\centering\arraybackslash}X}
  \scalebox{0.9}{
  \begin{tabularx}{\linewidth}{l C C C}
    \toprule
    \textbf{Method} & \textbf{\small LLaMA-3.2-3B} & \textbf{\small LLaMA-3.1-8B} & \textbf{\small Mistral-Nemo-12B} \\
    \midrule
    \midrule
    \multicolumn{4}{c}{\textbf{Keep Token Rate = 100\%}} \\
    \midrule
    FullKV & 44.23 & 49.32 & 48.32 \\
    \midrule
    \multicolumn{4}{c}{\textbf{Keep Token Rate = 10\%}} \\
    \midrule
    Oracle & 44.13 & 47.83 & 47.75 \\ \hline
    GemFilter & 35.20 & 37.59 & 36.60 \\
    FastKV & 42.45 & 46.81 & 45.62 \\
    SpecPrefill & 34.06 & 41.99 & N/A \\
    \rowcolor[HTML]{E8F8E8} CLAA & \textbf{42.93} & \textbf{47.13} & \textbf{46.05}  \\
    \midrule
    \multicolumn{4}{c}{\textbf{Keep Token Rate = 20\%}} \\
    \midrule
    Oracle & 44.35 & 48.39 & 48.27 \\ \hline
    GemFilter & 40.93 & 42.29 & 40.89 \\
    FastKV & 42.45 & 47.33 & \textbf{47.38} \\
    SpecPrefill & 35.90 & 44.57 & N/A \\
    \rowcolor[HTML]{E8F8E8} CLAA & \textbf{43.78} & \textbf{48.12} & 46.99 \\
    \midrule
    \multicolumn{4}{c}{\textbf{Keep Token Rate = 40\%}} \\
    \midrule
    Oracle & 44.52 & 48.88 & 48.24 \\ \hline
    GemFilter & 41.37 & 45.11 & 44.83 \\
    FastKV & \textbf{44.17} & 47.68 & \textbf{48.04} \\
    SpecPrefill & 39.17 & 46.55 & N/A \\
    \rowcolor[HTML]{E8F8E8} CLAA & 44.15 & \textbf{48.72} & 48.02 \\
    \bottomrule
  \end{tabularx}
  }
\end{table}

\begin{table}
\centering
\setlength{\tabcolsep}{3pt}
\renewcommand{\arraystretch}{1.1}
\small
\caption{RULER benchmark performance by category with 64K context on LLaMA-3.1-8B with 40\% token keep rate.}
\label{tab:ruler_results_categorized}
\begin{tabular}{l cccc | c}
\toprule
\textbf{Method} & \textbf{Retr.} & \textbf{M-Hop} & \textbf{Agg.} & \textbf{QA} & \textbf{Avg.} \\
\midrule
FullKV    & 98.27 & 86.88 & 86.67 & 63.30 & 83.78 \\
\midrule
GemFilter & 82.23 & 58.20 & \textbf{88.33} & 63.70 & 73.12 \\
FastKV    & 87.69 & 87.16 & 86.67 & 63.60 & 81.28 \\
\rowcolor[HTML]{E8F8E8} CLAA & \textbf{89.85} & \textbf{87.72} & 86.67 & \textbf{63.80} & \textbf{82.01} \\
\bottomrule
\end{tabular}
\end{table}

\subsection{The Cost of Causal Prediction}
\label{sec:oracle_gap}

Heuristic methods cannot match the Oracle because they lack information. The Oracle sees which prompt tokens the model attends to while generating the answer. Heuristics see only the prompt and must guess which tokens matter before generation begins. This information gap sets a ceiling on prefill compression: the difference between Oracle and heuristic performance is the cost of predicting without knowing the output. The Oracle also separates ranking quality from task difficulty. If a method scores poorly, we cannot tell whether the ranking is bad or the task is hard. The Oracle fixes a reference point, showing the minimum error given perfect information.

We expect this gap to vary by task structure. When the relevant tokens (e.g., keywords or entity names) are predictable from the prompt alone, heuristics should approach the Oracle. When the answer path depends on subtle contextual cues, the gap should widen. Token importance is not a static property of the prompt but depends on which output the model produces.

Our results confirm this pattern. On TriviaQA, heuristics nearly match the Oracle: CLAA achieves 92.37\% versus the Oracle's 91.43\% at 10\% keep rate, within noise. The query explicitly names the target entity, so heuristics identify relevant tokens without knowing the answer. In contrast, tasks like Qasper show larger gaps (CLAA 42.36\% vs. Oracle 43.94\%), suggesting the relevant context is harder to predict from the query alone.

The Oracle framework also clarifies the value of lookahead. Speculative Prefill generates tokens with a small model before ranking, partially closing the information gap. However, it still lags behind simpler heuristics: at 10\% keep rate, Speculative Prefill achieves 41.99\% versus CLAA's 47.13\% (Table~\ref{tab:longbench_avg_comparison}). The 1B speculator poorly predicts which tokens the 8B model will attend to, negating the benefit of lookahead.

\subsection{Efficiency and Accuracy Trade-off}
\label{sec:results:eff}

\begin{figure*}[t]
    \begin{minipage}[t]{0.37\linewidth}
        \centering
        \includegraphics[width=\linewidth]{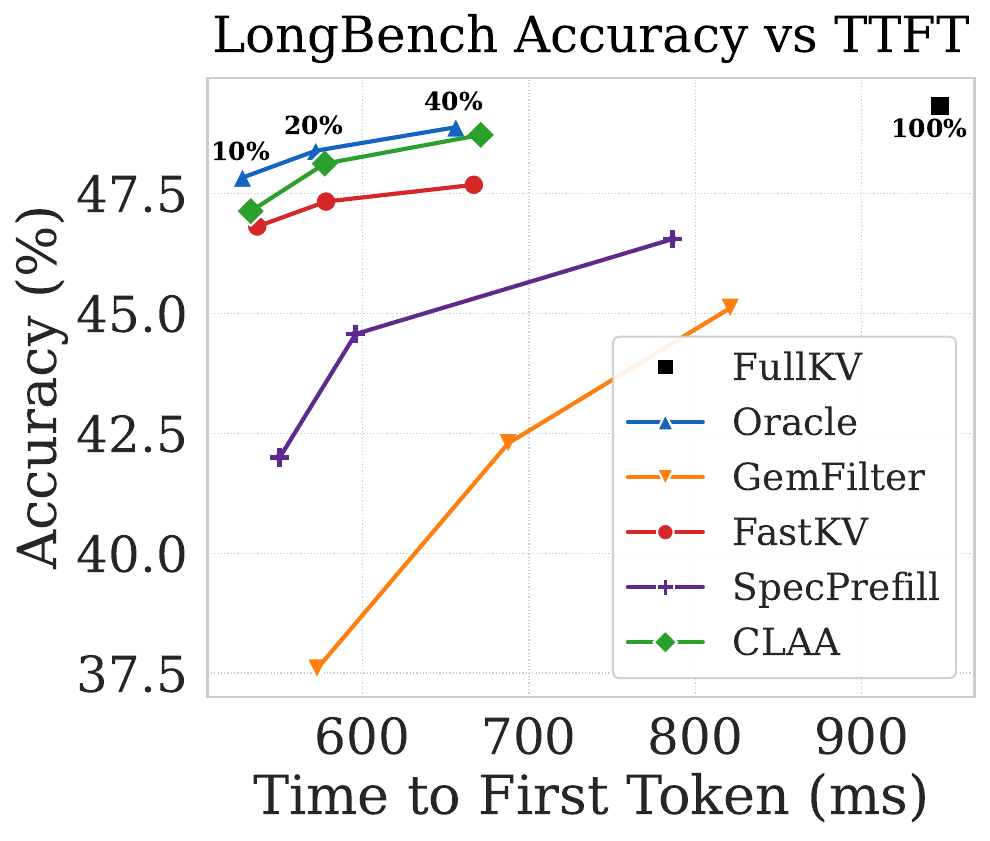}
        \caption{LongBench accuracy versus Time-to-First-Token (TTFT) for LLaMA-3.1-8B-Instruct on a 10k token sequence. Points correspond to 10\%, 20\%, and 40\% keep rates.}
        \label{fig:ttft-accuracy-analysis}
    \end{minipage}
    \hfill %
    \begin{minipage}[t]{0.61\linewidth}
        \centering
        \includegraphics[width=\linewidth]{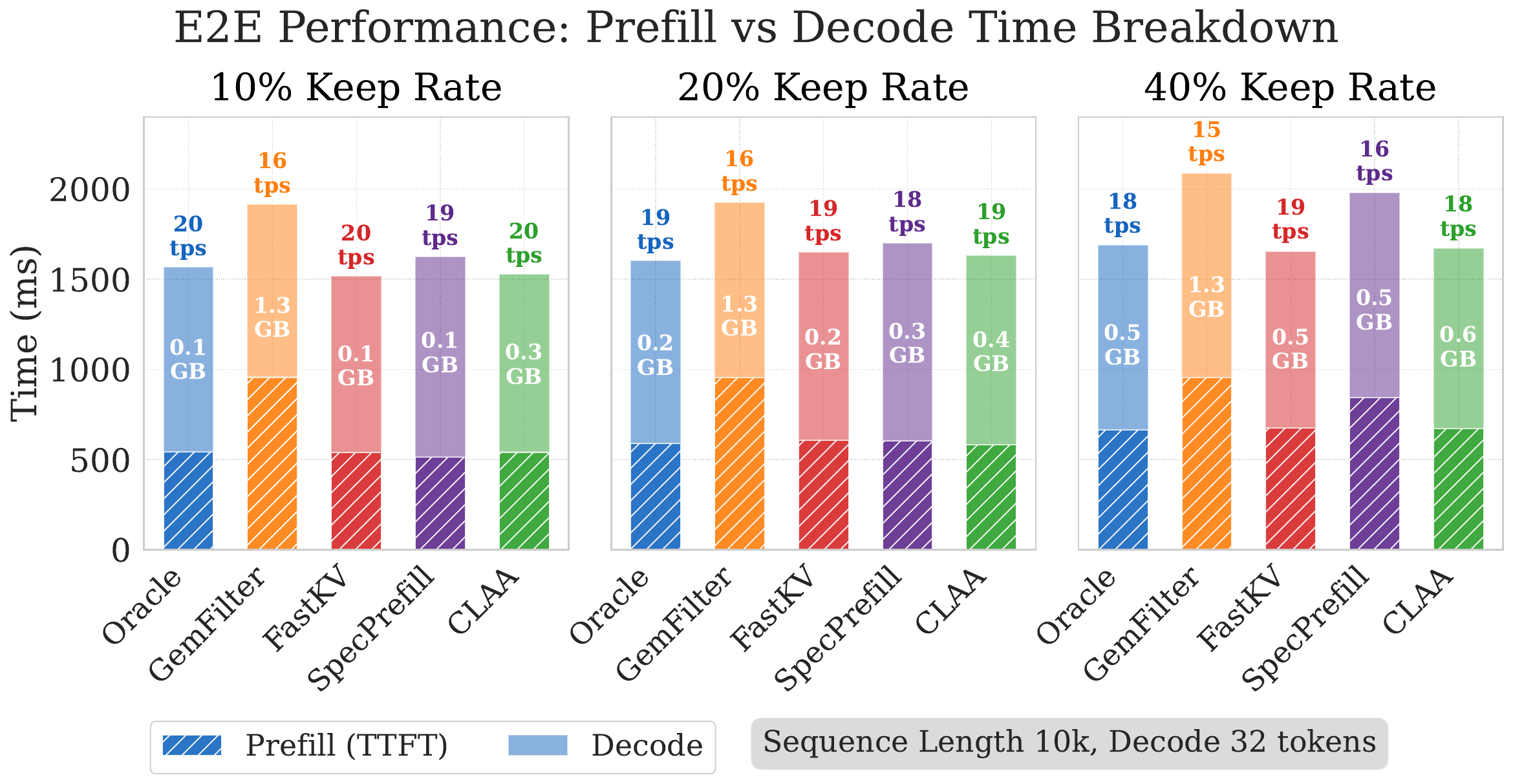}
        \caption{End-to-end performance breakdown for a 10k token prompt and 32 token generation. Bars show Prefill (TTFT) and Decode time. Annotations indicate decode throughput (tokens per second) and KV cache size (GB) at the start of decode.}
        \label{fig:e2e-performance}
    \end{minipage}
\end{figure*}

Figure~\ref{fig:ttft-accuracy-analysis} shows the accuracy-speed tradeoff for each method. CLAA tracks the Oracle ceiling more closely than single-layer methods, consistent with the diagnosis that layer instability was the limiting factor. At a 10\% keep rate, CLAA reduces TTFT by 39\% compared to FullKV (from roughly 900ms to 550ms). It achieves 47.13\% accuracy, closely approaching the oracle at 47.83\%. FastKV offers the next best performance. In contrast, GemFilter and Speculative Prefill exhibit a larger accuracy degradation for a similar reduction in TTFT. Figure~\ref{fig:e2e-performance} provides a detailed breakdown of prefill and decode performance. CLAA leaves the first four layers uncompressed to maintain foundational token representations before aggregation. This results in a minimal increase in KV cache size, for example, 0.3 GB for CLAA versus 0.1 GB for the Oracle at a 10\% keep rate. This modest increase in memory is justified by accuracy improvements (Figure~\ref{fig:ttft-accuracy-analysis}), showing a favorable trade-off between resource usage and performance. In contrast, the GemFilter approach results in a larger KV cache (1.3 GB at a 10\% keep rate) because its implementation retains the full cache and indexes into it during decoding, which negatively impacts decode throughput (16 tps compared to 19-20 tps for others).

\subsection{Comparison with Semantic Compression}
\label{sec:results:semantic}

Semantic compression methods take a different approach to prefill acceleration: they use auxiliary models to rewrite lengthy prompts into more concise representations before processing with the main model. We compare CLAA against two representative methods: Selective Context~\cite{li2023compressing} and LLMLingua~\cite{jiang2023llmlingua}. Both use LLaMA-3.2-1B-Instruct as the compression model and LLaMA-3.1-8B-Instruct as the main model.

Table~\ref{tab:semantic_compression} reports end-to-end TTFT including the compression overhead. The autoregressive rewrite step dominates total latency for both semantic compression methods. Selective Context is slower than the Full-KV baseline on all three tasks due to its 3--5 second rewrite overhead, while also degrading task quality. LLMLingua achieves modest speedups on longer contexts (GovReport) but still underperforms CLAA on both speed and quality. In contrast, CLAA achieves 1.34x--1.45x speedup over Full-KV while maintaining accuracy within 4\% of the baseline. Semantic compression methods could benefit in amortized settings where the same rewritten prompt is reused across multiple queries; however, for single-query inference, token-ranking heuristics like CLAA offer a more favorable efficiency-quality tradeoff.

\begin{table}[t]
\centering
\caption{Comparison with semantic compression methods on LLaMA-3.1-8B. Rewrite time is the compression model overhead; Total TTFT includes both rewrite and prefill.}
\label{tab:semantic_compression}
\renewcommand{\arraystretch}{1.1}
\scalebox{0.72}{
\begin{tabular}{llcccc}
\toprule
\textbf{Task} & \textbf{Method} & \textbf{Rewrite} & \textbf{Total TTFT} & \textbf{Speedup} & \textbf{Score} \\
\midrule
\multirow{4}{*}{Qasper}
  & Full KV & --- & 1084 ms & 1.00x & 0.487 \\
  & Selective Context & 3134 ms & 3421 ms & 0.32x & 0.186 \\
  & LLMLingua & 901 ms & 1141 ms & 0.95x & 0.129 \\
  & \cellcolor[HTML]{E8F8E8} CLAA (20\%) & \cellcolor[HTML]{E8F8E8} --- & \cellcolor[HTML]{E8F8E8} 755 ms & \cellcolor[HTML]{E8F8E8} \textbf{1.44x} & \cellcolor[HTML]{E8F8E8} \textbf{0.470} \\
\midrule
\multirow{4}{*}{GovReport}
  & Full KV & --- & 13505 ms & 1.00x & 0.377 \\
  & Selective Context & 5642 ms & 15391 ms & 0.88x & 0.327 \\
  & LLMLingua & 1815 ms & 9888 ms & 1.37x & 0.236 \\
  & \cellcolor[HTML]{E8F8E8} CLAA (20\%) & \cellcolor[HTML]{E8F8E8} --- & \cellcolor[HTML]{E8F8E8} 10055 ms & \cellcolor[HTML]{E8F8E8} \textbf{1.34x} & \cellcolor[HTML]{E8F8E8} \textbf{0.310} \\
\midrule
\multirow{4}{*}{TriviaQA}
  & Full KV & --- & 1768 ms & 1.00x & 0.944 \\
  & Selective Context & 5254 ms & 6027 ms & 0.29x & 0.204 \\
  & LLMLingua & 1682 ms & 2412 ms & 0.73x & 0.470 \\
  & \cellcolor[HTML]{E8F8E8} CLAA (20\%) & \cellcolor[HTML]{E8F8E8} --- & \cellcolor[HTML]{E8F8E8} 1217 ms & \cellcolor[HTML]{E8F8E8} \textbf{1.45x} & \cellcolor[HTML]{E8F8E8} \textbf{0.944} \\
\bottomrule
\end{tabular}
}
\end{table}

\subsection{Ablation Studies}
\label{sec:results:ablations}

\begin{figure}[t]
    \centering
    \includegraphics[width=\linewidth]{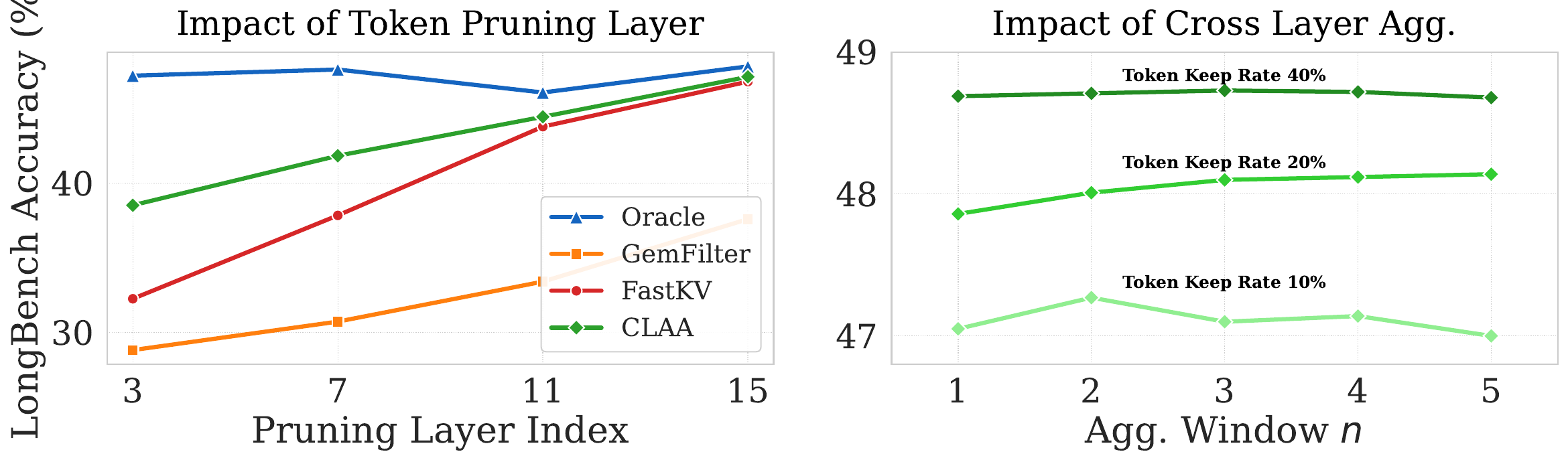}
    \caption{\textbf{Left}: Impact of the pruning layer index on LongBench accuracy. Pruning later in the model improves performance for all heuristics. \textbf{Right}: Impact of the cross-layer aggregation window size ($n$) at different token keep rates. Increasing the window size improves accuracy.}
    \label{fig:ablations}
\end{figure}

\paragraph{Impact of Pruning Layer.}
Figure~\ref{fig:ablations} (left) demonstrates a strong positive correlation between LongBench accuracy and the pruning layer index. For all heuristics, applying pruning at later layers results in higher accuracy. For instance, the accuracy of FastKV increases from 32.5\% at layer 3 to 46.5\% at layer 15. This finding indicates that premature pruning at early layers, which compute foundational representations, leads to information loss. CLAA consistently outperforms both FastKV and GemFilter across all tested layers, particularly at earlier ones, which shows the robustness of cross-layer aggregation. As the pruning layer index increases, the performance of all methods improves and converges toward the upper bound established by the oracle. The selection of layer 15 for our main experiments, positioned approximately halfway through the model, is therefore justified as a balance between computational savings and performance.

\paragraph{Impact of Cross-Layer Aggregation Window.}
Figure~\ref{fig:ablations} (right) illustrates the effect of the aggregation window size, $n$, on the performance of CLAA at different token keep rates. For all keep rates, expanding the window from $n=1$ to $n=2$ provides an accuracy boost, validating the core hypothesis that cross-layer aggregation mitigates the instability of single-layer ranking. Larger windows have varied effects, with performance for higher token keep rates (20\% and 40\%) typically stabilizes around $n=4$. For the highly aggressive 10\% keep rate, performance is more volatile, peaking at $n=2$. These results demonstrate that a small aggregation window is sufficient to stabilize rankings. A window size of $n=4$ was selected for our experiments as it offers a robust performance benefit across various keep rates with minimal overhead.

\section{Limitations}
\label{sec:limitations}

While CLAA achieves strong performance across retrieval and QA tasks, several limitations warrant discussion. First, summarization tasks such as Multi-News exhibit relatively flat rank correlations for all methods, including CLAA (Figure~\ref{fig:ranking-comparison}). In these tasks, token importance appears to evolve dynamically throughout generation rather than being fixed at the prompt boundary. Static prefill pruning, which commits to a single importance ranking before any tokens are generated, may be fundamentally limited for such workloads. Future methods may need to re-assess token importance dynamically during decoding to capture these shifts.

Second, our evaluation focuses on single-turn inference. In multi-turn conversation settings, token pruning methods face an additional challenge: importance estimates from the first user turn may not remain valid as the conversation evolves. While these methods could theoretically be re-run for each additional turn, doing so would eventually negate the efficiency gains. Extending token-ranking heuristics to multi-turn settings remains an open problem shared by all methods in this class.

Finally, all methods in our evaluation, including CLAA, use FlashAttention-2 for the main forward pass. The scoring computation required by CLAA (an $W \times L$ attention computation per layer in the aggregation window) accounts for less than 2\% of total TTFT and is already included in our reported wall-clock times. No modifications to underlying CUDA kernels are required.

\section{Conclusion}
We introduce the Answer-Informed Oracle, a framework that defines ground-truth token importance for evaluating prefill acceleration heuristics. The oracle reveals that existing methods suffer from layer-wise ranking instability, a failure mode invisible to end-to-end benchmarks. This diagnosis suggests a simple fix: aggregate across layers. CLAA implements this idea and closes the gap to the oracle ceiling while reducing Time-to-First-Token by up to 39\%.

\newpage
\section*{Impact Statement}
This paper presents work whose goal is to advance the field of machine learning. There are many potential societal consequences of our work, none of which we feel must be specifically highlighted here.

\bibliography{references}
\bibliographystyle{icml2026}

\appendix

\section{Hyperparameter Configuration}
\label{app:hyperparameters}

This appendix details the hyperparameter configurations for all methods to ensure a transparent and reproducible comparison. Key architectural parameters, such as the pruning layer index, were swept across all applicable methods for our ablation studies (Section \ref{sec:results:ablations}). The final values used in our main results (e.g., Tables \ref{tab:longbench_results_generated} and \ref{tab:longbench_avg_comparison}) were selected to provide a consistent and fair comparison point across methods.

\begin{table*}[!ht]
\centering
\caption{Hyperparameter settings for all methods evaluated on LLaMA-3.1-8B-Instruct.}
\label{tab:hyperparameters}
\renewcommand{\arraystretch}{1.1}
\scalebox{0.85}{
\begin{tabular}{@{}llcll@{}}
\toprule
\textbf{Method} & \textbf{Parameter} & \textbf{Symbol} & \textbf{Value(s) Explored} & \textbf{Final Value in Main Results} \\
\midrule
\textbf{Oracle} & Token Keep Rate & - & \{0.1, 0.2, 0.4\} & Main Variable \\
 (Ours)           & Pruning Layer Index & $l_p$ & \{3, 7, 11, 15, 19\} & 15 \\
\midrule
\textbf{FastKV} & Token Keep Rate & - & \{0.1, 0.2, 0.4\} & Main Variable \\
            & TSP Layer Index & $l_{\text{TSP}}$ & \{3, 7, 11, 15, 19\} & 15 \\
            & Observation Window & $W$ & \{8\} & 8 \\
            & Pooling Kernel Size & - & \{7\} & 7 \\
\midrule
\textbf{GemFilter} & Token Keep Rate & - & \{0.1, 0.2, 0.4\} & Main Variable \\
            & Routing Layer Index & $r$ & \{3, 7, 11, 15, 19\} & 15 \\
\midrule
\textbf{SpecPrefill} & Token Keep Rate & - & \{0.1, 0.2, 0.4\} & Main Variable \\
            & Lookahead Tokens & $k$ & \{8\} & 8 \\
\midrule
\textbf{CLAA} & Token Keep Rate & - & \{0.1, 0.2, 0.4\} & Main Variable \\
(Ours)      & Pruning Layer Index & $l_p$ & \{3, 7, 11, 15, 19\} & 15 \\
            & Aggregation Window & $n$ & \{1, 2, 4, 8, 12\} & 4 \\
            & First Uncompressed Layer & $m$ & \{0, 2, 4, 6\} & 4 \\
            & Observation Window & $W$ & \{8\} & 8 \\
            & Pooling Kernel Size & - & \{7\} & 7  \\
\bottomrule
\end{tabular}
}
\end{table*}

\section{Ablation on First Compression Layer (\textit{m})}
\label{app:ablation_m}

\begin{figure}[!ht]
    \centering
    \includegraphics[width=0.5\columnwidth]{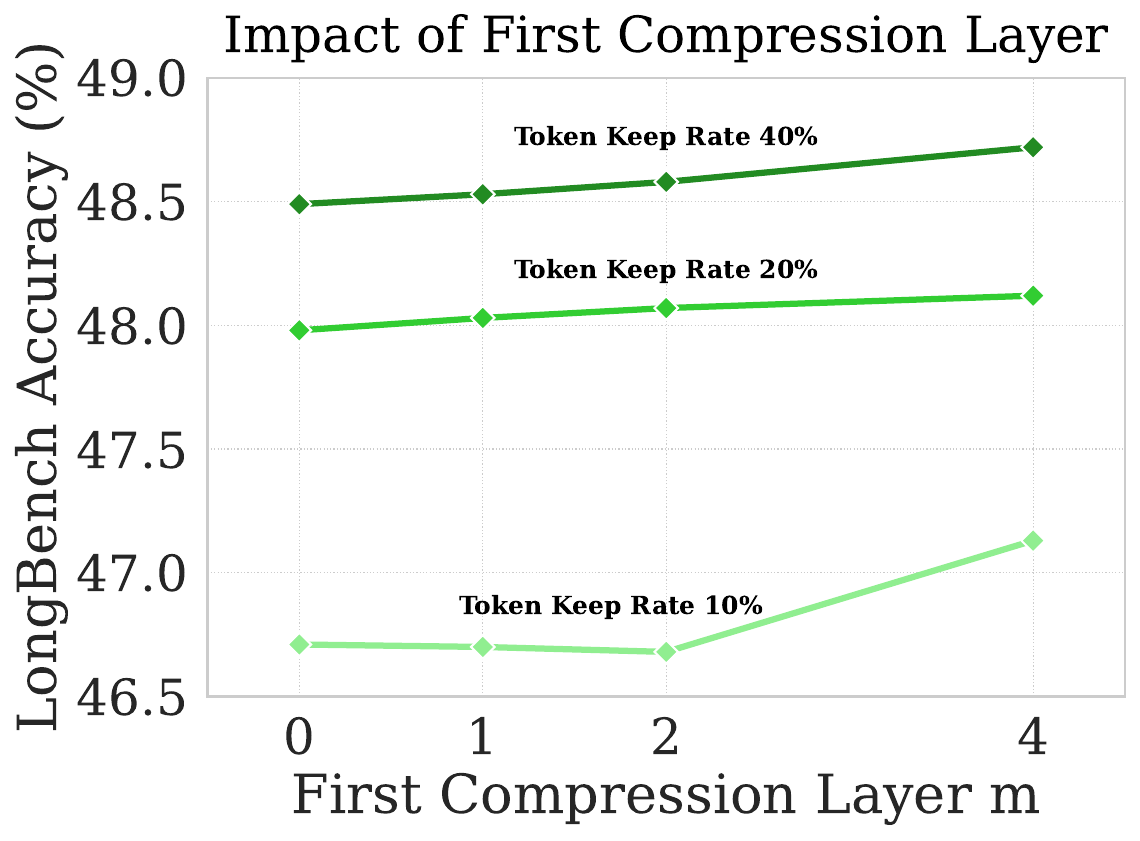}
    \caption{Impact of the first compression layer index ($m$) on LongBench accuracy for CLAA. For $m>0$, layers $0$ to $m-1$ are kept uncompressed. The experiment shows that deferring compression to later layers (increasing $m$) improves accuracy, especially for higher token keep rates. These results are on Llama-3.1-8B-Instruct.}
    \label{fig:ablation_m}
\end{figure}

In our main experiments, we set the number of initial uncompressed layers for CLAA to $m=4$. This decision was based on an ablation study investigating the trade-off between early-layer compression and model accuracy. As shown in Figure~\ref{fig:ablation_m}, deferring compression to later layers consistently improves performance across all token keep rates. Starting compression from layer 0 (i.e., $m=0$) results in the lowest accuracy, as it forces the model to build its initial representations from a compressed context. Delaying compression until layer 4 provides an accuracy boost, justifying the choice of $m=4$ as a default that balances performance and computational efficiency.

\section{Per-Task Ranking Correlation Analysis}
\label{app:per_task_analysis}

\begin{figure*}
    \centering
    \includegraphics[width=0.86\textwidth]{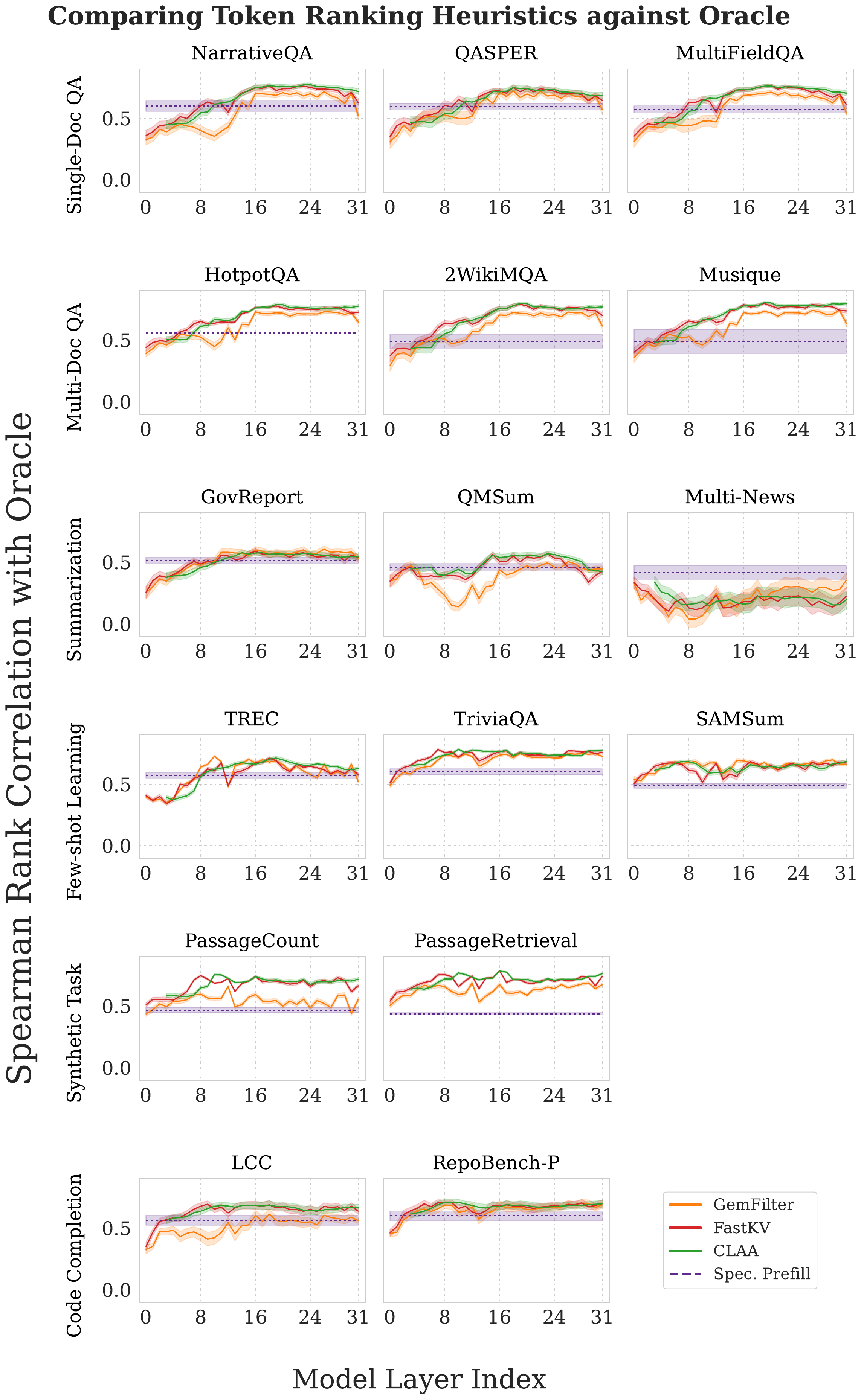}
    \caption{Detailed per-task comparison of token ranking heuristics against the Answer-Informed Oracle.}
    \label{fig:appendix_ranking}
\end{figure*}

This section analyzes token ranking performance by comparing various heuristics against the Answer-Informed Oracle. Figure~\ref{fig:appendix_ranking} shows Spearman Rank Correlation for each heuristic across all model layers.

\begin{itemize}
   \item \textbf{Layer-specific volatility:} Single-layer heuristics show performance drops at specific layers. In NarrativeQA, GemFilter and FastKV exhibit sharp correlation drops around layer 10. This pattern repeats across QASPER, HotpotQA, and QMSum. CLAA avoids these drops through multi-layer aggregation.

   \item \textbf{Practical implications:} Methods using single pre-selected layers risk severe performance degradation if their chosen layer coincides with a performance trough. CLAA provides more reliable token selection by aggregating across a layer window.

   \item \textbf{Performance convergence:} While layer-dependent heuristics reach similar high performance in deep layers (20-31), CLAA achieves comparable results without the intermediate volatility.

   \item \textbf{Universal patterns:} Early layers (0-8) consistently fail to predict token importance across all methods. Speculative Prefill serves as a task-dependent baseline with varying competitiveness.
\end{itemize}

These results demonstrate that CLAA provides more stable token ranking compared to single-layer methods, making it a robust choice for prefill acceleration.

\section{Token Ranking Heuristic Pseudocode}
\label{app:pseudocode}

This section provides detailed and unified pseudocode for the token ranking heuristics discussed in the paper: GemFilter, FastKV, Speculative Prefill, our Answer-Informed Oracle, and our proposed method, CLAA.

\subsection{GemFilter}

GemFilter operates in a two-pass system. It first runs a partial forward pass to rank tokens using the query from the final prompt token. It then discards this intermediate state and executes a new, standard forward pass using only the top-ranked tokens and their original positions. This architectural approach is outlined in Listing~\ref{lst:gemfilter}.

\subsection{FastKV}

Unlike GemFilter's two-pass system, FastKV operates in a single, continuous forward pass. At each layer, it uses an ``observation window'' of recent tokens to rank all prompt tokens and compress the stored KV cache. At a designated Token-Selective Propagation (TSP) layer, it performs a one-time sequence pruning, after which the forward pass continues on the reduced sequence. This integrated architecture is outlined in Listing~\ref{lst:fastkv}.

\subsection{Speculative Prefill}

Speculative Prefill uses a two-model architecture. A small ``speculator'' model first performs a full prefill and generates a few lookahead tokens. The queries from these generated tokens are used to rank the importance of the original prompt tokens. Finally, the large ``base'' model performs a selective prefill on only the top-ranked prompt tokens, using their original position IDs to maintain context. This multi-stage process is outlined in Listing~\ref{lst:spec_prefill}.

\subsection{Answer-Informed Oracle}

The Answer-Informed Oracle provides a ground-truth token ranking by leveraging knowledge of the final generated answer. Its operation is split into two phases. First, in an offline ranking phase, it generates a complete answer and measures the attention from each generated token back to the original prompt tokens. Second, in an online evaluation phase, this pre-computed ranking is used to perform a selective prefill, establishing a theoretical performance upper bound. This process is detailed in Listing~\ref{lst:oracle}.

\subsection{Cross-Layer Attention Aggregation (CLAA)}

Our proposed method, CLAA, enhances the single-pass architecture by improving ranking stability. It first defers any compression for an initial $m$ layers. Then, for each subsequent layer, it computes importance scores using an observation window and stores them in a buffer of size $n$. At a designated pruning layer, $l_p$, it aggregates scores across this buffer by taking the maximum value for each token. This process is outlined in Listing~\ref{lst:claa}.

\begin{listing*}[t!]
\begin{lstlisting}[language=Python]
def gemfilter_prefill(model, prompt_tokens, routing_layer, keep_rate):
    # Run a partial forward pass to get the query from the last prompt token
    hidden_states = model.forward(prompt_tokens, stop_at_layer=routing_layer)
    q_last = hidden_states.get_last_token_query(layer=routing_layer)
    k_all = hidden_states.get_keys(layer=routing_layer)

    # Rank tokens by their raw attention (pre-softmax) to the last token q
    scores = aggregate_attention(q_last, k_all, use_softmax=False)
    top_k_indices = topk(scores, keep_rate).indices


    pruned_tokens = gather(prompt_tokens, top_k_indices)

    # Restore position ids
    original_position_ids = gather(range(len(prompt_tokens)), top_k_indices)

    # Execute a full forward pass on pruned sequence
    final_outputs, final_kv_cache = model.forward(
        pruned_tokens,
        position_ids=original_position_ids
    )

    return final_outputs, final_kv_cache
\end{lstlisting}
    \caption{GemFilter Prefill Logic.}
    \label{lst:gemfilter}
\end{listing*}

\begin{listing*}
\begin{lstlisting}[language=Python]
def fastkv_prefill(model, prompt_tokens, tsp_layer, window_size, keep_rate):
    hidden_state = model.embed(prompt_tokens)
    kv_cache = {}

    for l in range(model.num_layers):
        # Get queries from the observation window
        q_window = hidden_state.get_last_n_queries(n=window_size, layer=l)
        k_all = hidden_state.get_keys(layer=l)
        v_all = hidden_state.get_values(layer=l)

        # Rank tokens using post-softmax attention from the window
        scores = aggregate_attention(q_window, k_all, use_softmax=True)
        top_k_indices = topk(scores, keep_rate).indices

        # Compress KV cache using rankings (for decode)
        kv_cache[l] = gather(k_all, v_all, on_indices=top_k_indices)

        hidden_state = model.layer_forward(l, hidden_state, use_kv=(k_all, v_all))

        # At the TSP layer, prune hidden state with rankings
        if l == tsp_layer:
            hidden_state = gather(hidden_state, on_indices=top_k_indices)

    final_outputs = model.final_norm(hidden_state)
    return final_outputs, kv_cache
\end{lstlisting}
    \caption{FastKV Prefill Logic.}
    \label{lst:fastkv}
\end{listing*}

\begin{listing*}
\begin{lstlisting}[language=Python]
def speculative_prefill(base_model, spec_model, prompt_tokens, look_ahead_k, keep_rate):
    # Run a small speculator model on the full prompt
    spec_outputs, spec_kv_cache = spec_model.forward(prompt_tokens)
    spec_k_prompt = spec_kv_cache.get_all_keys()

    # Generate `look_ahead_k` tokens with the speculator to get qs
    q_generated = []
    next_token = spec_outputs.get_next_token()
    for _ in range(look_ahead_k):
        lookahead_out, spec_kv_cache = spec_model.forward(next_token, spec_kv_cache)
        q_generated.append(lookahead_out.get_query())
        next_token = lookahead_out.get_next_token()

    # Rank prompt tokens using draft qs
    scores = aggregate_attention(q_generated, spec_k_prompt)
    top_k_indices = topk(scores, keep_rate).indices

    pruned_tokens = gather(prompt_tokens, top_k_indices)
    original_position_ids = gather(range(len(prompt_tokens)), top_k_indices)
    final_outputs, final_kv_cache = base_model.forward(
        pruned_tokens,
        position_ids=original_position_ids
    )
    return final_outputs, final_kv_cache
\end{lstlisting}
    \caption{Speculative Prefill Logic.}
    \label{lst:spec_prefill}
\end{listing*}

\begin{listing*}
\begin{lstlisting}[language=Python]
def get_oracle_ranking(model, prompt_tokens, max_gen_len):
    _, prefill_kv_cache = model.forward(prompt_tokens)
    k_prompt = prefill_kv_cache.get_all_keys()

    q_generated = []
    next_token = prefill_kv_cache.get_next_token()
    current_kv_cache = prefill_kv_cache
    while not is_eos(next_token) and len(q_generated) < max_gen_len:
        output, current_kv_cache = model.forward(next_token, past_kv=current_kv_cache)
        q_generated.append(output.get_query())
        next_token = output.get_next_token()

    scores = aggregate_attention(q_generated, k_prompt, use_softmax=False)

    return scores

def oracle_prefill(model, prompt_tokens, oracle_scores, keep_rate):
    # Use the pre-computed oracle scores to select the top-k indices
    top_k_indices = topk(oracle_scores, keep_rate).indices

    pruned_tokens = gather(prompt_tokens, top_k_indices)
    original_position_ids = gather(range(len(prompt_tokens)), top_k_indices)

    final_outputs, final_kv_cache = model.forward(
        pruned_tokens,
        position_ids=original_position_ids
    )
    return final_outputs, final_kv_cache
\end{lstlisting}
    \caption{Answer-Informed Oracle Prefill Logic.}
    \label{lst:oracle}
\end{listing*}

\begin{listing*}
\begin{lstlisting}[language=Python]
def claa_prefill(model, prompt_tokens, pruning_layer, aggregation_window,
                 defer_layers, window_size, keep_rate):
    hidden_state = model.embed(prompt_tokens)
    kv_cache = {}
    layer_scores_buffer = collections.deque(maxlen=aggregation_window)

    for l in range(model.num_layers):
        # Defer any compression for the first `defer_layers`
        if l < defer_layers:
            hidden_state = model.layer_forward(l, hidden_state)
            kv_cache[l] = hidden_state.get_kv_pair()
            continue

        # Get queries from the observation window
        q_window = hidden_state.get_last_n_queries(n=window_size, layer=l)
        k_all, v_all = hidden_state.get_keys(layer=l), hidden_state.get_values(layer=l)

        # Compute importance scores for the current layer using the observation window
        current_layer_scores = aggregate_attention(q_window, k_all, use_softmax=True)
        # Store the scores in a rolling buffer for later aggregation.
        layer_scores_buffer.append(current_layer_scores)

        # Compress KV cache using rankings (for decode)
        compression_indices = topk(current_layer_scores, keep_rate).indices
        kv_cache[l] = gather(k_all, v_all, on_indices=compression_indices)

        hidden_state = model.layer_forward(l, hidden_state, use_kv=(k_all, v_all))

        # At the TSP layer, prune hidden state with agg. rankings
        if l == pruning_layer:
            # Aggregate scores (Eq. 6 in paper)
            aggregated_scores = max(layer_scores_buffer, dim=0)
            pruning_indices = topk(aggregated_scores, keep_rate).indices
            hidden_state = gather(hidden_state, on_indices=pruning_indices)

    final_outputs = model.final_norm(hidden_state)
    return final_outputs, kv_cache
\end{lstlisting}
    \caption{Cross-Layer Attention Aggregation (CLAA) Prefill Logic.}
    \label{lst:claa}
\end{listing*}

\end{document}